\documentclass[11pt]{article}
\usepackage{amsfonts}

\usepackage[final]{acl}
\usepackage{makecell}
\newcommand{\best}[1]{{\bfseries\tablenum{#1}}}
\usepackage{listings}
\usepackage{xcolor}
\usepackage[table]{xcolor}
\usepackage{booktabs}
\usepackage{multirow}
\usepackage{makecell}
\usepackage{siunitx} 
\usepackage{times}
\usepackage{latexsym}
\usepackage{xcolor}
\usepackage{pifont}
\usepackage{amsmath}
\usepackage[most]{tcolorbox}
\usepackage{adjustbox}
\newcommand{\cmark}{\textcolor{green!60!black}{\ding{51}}} 
\newcommand{\xmark}{\textcolor{red!70!black}{\ding{55}}}   

\lstdefinestyle{prompttext}{
  basicstyle=\ttfamily\small,
  columns=fullflexible,
  breaklines=true,
  breakatwhitespace=true,
  frame=none,
  showstringspaces=false
}

\newtcolorbox{promptbox}[1]{
  colback=white,
  colframe=black,
  boxrule=0.4pt,
  arc=1.5mm,
  left=1mm,right=1mm,top=1mm,bottom=1mm,
  title={#1},
  fonttitle=\bfseries
}

\usepackage[T1]{fontenc}

\usepackage[utf8]{inputenc}

\usepackage{microtype}
\usepackage[most]{tcolorbox}
\usepackage{enumitem}
\usepackage{inconsolata}

\usepackage{graphicx}

\title{Making Clinical Language Models Auditable: Concept-Guided Fine-Tuning for Robust Prediction}

\author{Jin Mu \\
  University of Wisconsin--Madison \\
  \texttt{jmu27@wisc.edu} \\\And
  Guanhua Chen \\
University of Wisconsin--Madison \\
  \texttt{gchen25@wisc.edu} \\}

\begin{document}
\maketitle
\begin{abstract}
Clinical language models can achieve strong in-hospital accuracy yet fail under
deployment shifts because they exploit note-specific artifacts (e.g., templates,
separators, boilerplate) that do not reflect patient state. We propose \textbf{CAST
(Concept-guided Artifact Suppression Tuning)}, an SAE-based framework for auditable
clinical text classification. CAST uses Sparse Autoencoders to expose sparse,
human-auditable features from intermediate Transformer activations, labels SAE
latents with an LLM-assisted interpretation pipeline and ICD-10 retrieval
constraints, suppresses verified artifact latents via residual subtraction during
fine-tuning, and provides post-hoc per-concept attributions for auditing model
decisions. On MIMIC-IV discharge-note mortality prediction, CAST improves over
its corresponding fine-tuned encoder baselines and remains competitive with strong
LLM baselines, while producing a feature-level audit trail of the clinical concepts
that support each prediction and the artifact concepts suppressed during training.
\end{abstract}

\section{Introduction}
\begin{figure*}[t]
  
  \centering
  \includegraphics[width=0.9\linewidth]{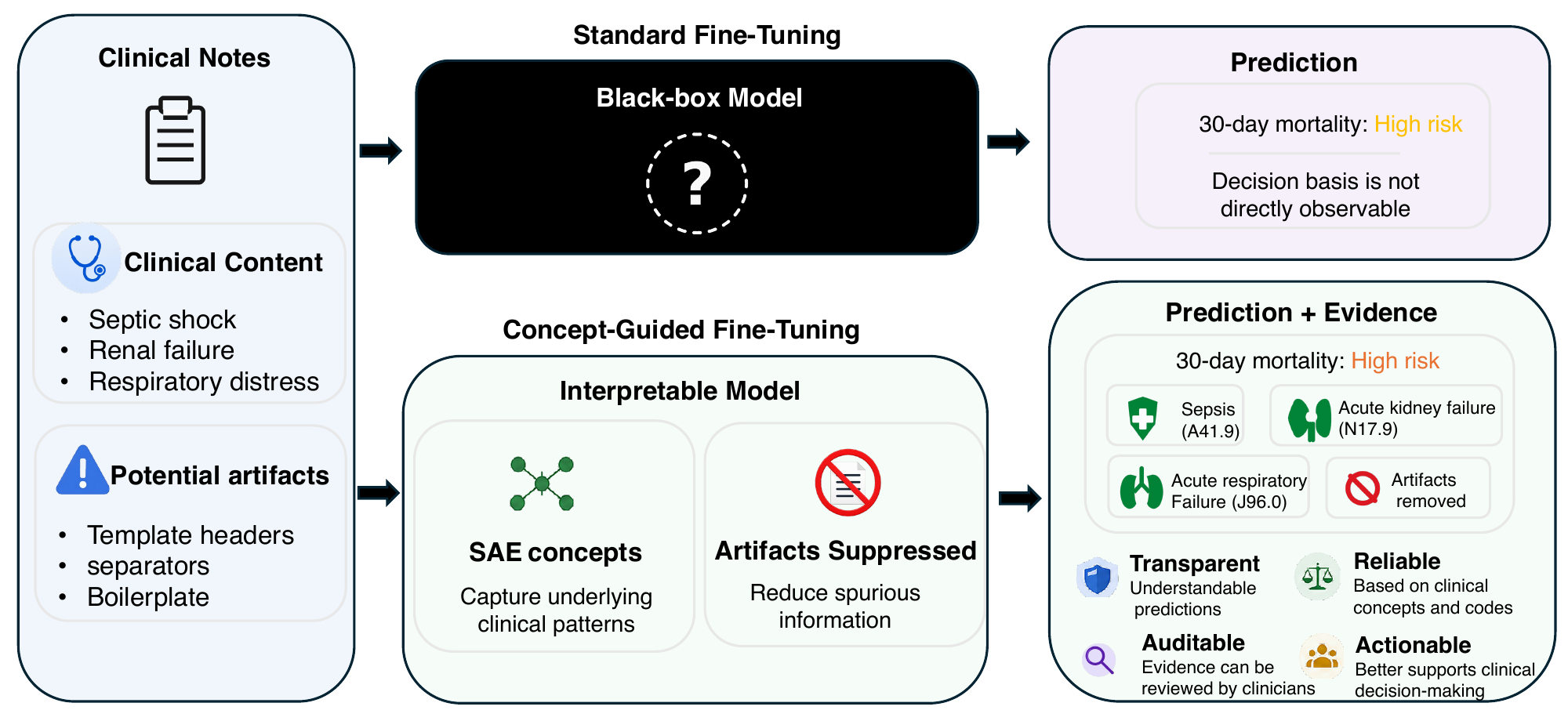} 
  \caption {\textbf{SAE-Guided Fine-Tuning Enables Transparent Clinical Risk Prediction.} Unlike standard fine-tuning, which yields opaque risk predictions, our framework encourages models to rely on clinically meaningful SAE concepts and suppress non-clinical artifacts, producing predictions with interpretable clinical evidence and ICD-10 codes.}
  \label{fig1}
\end{figure*}
Clinical language models have shown significant potential to transform clinical workflows by processing unstructured electronic health record (EHR) data for tasks such as clinical summarization, diagnostic reasoning, and risk prediction \citep{maity2025large,MENG2024109713,thirunavukarasu2023large}. Compared with structured variables alone, free-text clinical notes contain rich longitudinal information about patient history, disease progression, clinician assessments, and treatment decisions. This makes them especially valuable for high-acuity prediction tasks such as mortality risk estimation. However, despite strong in-distribution performance, the deployment of clinical
LLMs remains limited in high-stakes medical settings because their predictions
are often difficult to interpret, validate, and trust
\citep{amann2020explainability,markus2021role}. A central obstacle is shortcut
learning: models trained on EHR notes may exploit artifact-based signals that
are highly predictive within a dataset but clinically meaningless or unstable
under deployment shift. These signals can include note templates, section
headers, documentation style, repeated boilerplate, separators, discharge
formatting, or institution-specific coding patterns, which may correlate with
outcomes because of local documentation practices rather than true patient
physiology \citep{geirhos2020shortcut}.

This concern is especially acute in Intensive Care Unit (ICU) settings, where
risk predictions may inform time-sensitive decisions. Even when a model
correctly identifies a high-risk patient, the prediction has limited clinical
value if it is driven by such shortcuts rather than evidence of clinical
deterioration. Clinicians cannot justify decisions such as escalating life
support or initiating palliative care based on an opaque alert. Clinical
interpretability therefore requires more than highlighting salient words or
spans: it must expose the internal concepts that causally influence predictions
and determine whether they reflect medically meaningful evidence
\citep{rudin2019stop}.

Interpretability for clinical language models has traditionally focused on token-level feature attribution, employing label-wise attention or post-hoc saliency methods such as Integrated Gradients, LIME, and SHAP \citep{mullenbach2018explainablepredictionmedicalcodes, 10.5555/3491440.3491901,kim2022anemic,dolk2022evaluation}. While these techniques identify salient input spans, they are mechanistically limited: they pinpoint where a model attends without elucidating the underlying concepts or the functional logic governing their processing. Furthermore, the reliability of such attributions is frequently contested; prior work has demonstrated that attention weights can be poorly correlated with model outputs, posing significant risks for high-stakes clinical auditing \citep{jain2019attention}. Consequently, there remains a critical gap in moving beyond superficial, input-level correlations toward understanding the internal representations that drive clinical reasoning.

Recent work in mechanistic interpretability seeks to reverse-engineer models by decomposing internal neural activations into distinct, human-understandable features \citep{elhage2021mathematical,olsson2022context}. Specifically, Sparse Autoencoders (SAEs) have proven highly effective in general-domain models by disentangling polysemantic neurons into distinct "monosemantic" features \cite{bricken2023towards,gao2024scaling}. While SAEs improve model transparency, their application in the medical domain remains relatively limited. Prior medical-domain work has used dictionary features for mechanistic explanations and demonstrated feature-based steering \citep{wu-etal-2024-beyond}, but integrating such interpretable features as explicit artifact-suppression interventions during task-specific fine-tuning remains underexplored.

In this work, we propose \textbf{CAST (Concept-guided Artifact Suppression Tuning)}, a fine-tuning framework for clinical classification that transforms mechanistic interpretability from a passive analytical tool into an active steering mechanism. As illustrated in Figure~\ref{fig1}, standard fine-tuning maps clinical notes directly to risk predictions through a black-box model, making it difficult to determine whether predictions are based on meaningful clinical evidence or dataset-specific artifacts. In contrast, our framework uses Sparse Autoencoders trained on the large-scale MIMIC datasets \citep{johnson2016mimiciii,johnson2016mimic,PhysioNet-mimiciv-2.2} to identify internal model features corresponding to human-understandable clinical concepts. We then use these semantic units to guide fine-tuning by suppressing features associated with spurious shortcuts, such as formatting artifacts, while encouraging reliance on clinically meaningful representations. We evaluate our approach on MIMIC-IV ICU discharge notes for 30-day out-of-hospital mortality prediction and show that integrating concept-level control into the training loop yields not only improved predictive performance, but also a transparent feature-level audit trail with supporting clinical concepts and ICD-10 codes.

\section{Related Work}
\paragraph{Feature Extraction via Sparse Autoencoders}
To optimize the extraction of monosemantic features, several SAE architectures have emerged. Vanilla SAEs isolate monosemantic features by projecting dense activations into a high-dimensional, sparse latent space optimized with an $\ell_1$ penalty \cite{ng_sparse_autoencoder}. To overcome the feature shrinkage inherent to $\ell_1$ regularization, TopK SAEs enforce a hard activation budget by retaining only the K largest values per input \cite{gao2024scaling}. The BatchTopK variant builds on this by applying selection across entire batches to prevent dead neurons and stabilize training \cite{bussmann2024batchtopk}. Additionally, architectures like Matryoshka SAE explore hierarchical representations by constructing nested dictionaries, dividing the latent space into expanding prefixes to encode high-level abstractions in early dimensions and fine-grained details in later ones \cite{pmlr-v267-bussmann25a}. Together, these developments have substantially improved the quality, stability, and semantic organization of SAE-derived features, making them increasingly useful for downstream interpretability and intervention.

\paragraph{SAE-Guided Control and Adaptation}
Once high-quality monosemantic features are extracted, the interpretability
paradigm increasingly shifts from passive analysis toward active model intervention.
At inference time, several frameworks utilize SAE features for causal
interventions---systematically amplifying or ablating specific latents to reliably
steer model generation \citep{bayat2025steering}. This steering
capability has also been extended to complex cognitive and in-context mechanisms;
for instance, \citet{chen2025does} combine SAEs with activation patching to isolate
``reasoning features'' and probe Chain-of-Thought (CoT) faithfulness, while
\citet{cho2025toward} use SAE-guided procedures to track information flow and
improve in-context learning behavior. Beyond inference-time steering, recent work
has begun incorporating SAE-derived features into model adaptation. For example,
\citet{casademunt2025steering} integrate SAE-based concept ablation into fine-tuning
to suppress unintended generalizations such as gender bias, while Self-Regul
\citep{wu2025self} uses sparse autoencoders to regularize LLM-based classification
toward interpretable sparse features. However, these methods primarily target
general-domain generation, reasoning, behavioral steering, or controllable
classification, rather than clinical risk prediction. Our work addresses this gap
by adapting SAE-guided learning to clinical prediction, where the goal is not only
to steer model behavior but also to suppress clinically meaningless documentation
artifacts.

\paragraph{Mechanistic Interpretability in Clinical Classification}
Mechanistic interpretability has recently been explored as a way to expose and manipulate internal representations for text classification. In general-domain settings, SPIN \cite{jiao-etal-2024-spin} identifies and integrates task-relevant internal neurons to obtain more compact and interpretable classifiers.   \citet{gallifant-etal-2025-sparse} show that features discovered by Sparse Autoencoders can serve as effective classifier representations and transfer across models and modalities. However, these approaches do not use clinically interpreted artifact concepts as explicit suppression targets during task-specific fine-tuning. CAST instead integrates SAE-derived concepts directly into clinical model adaptation, updating representations during training to suppress spurious documentation artifacts.

\section{Method}
\begin{figure*}[t]
  
  \centering
  \includegraphics[width=0.88\linewidth]{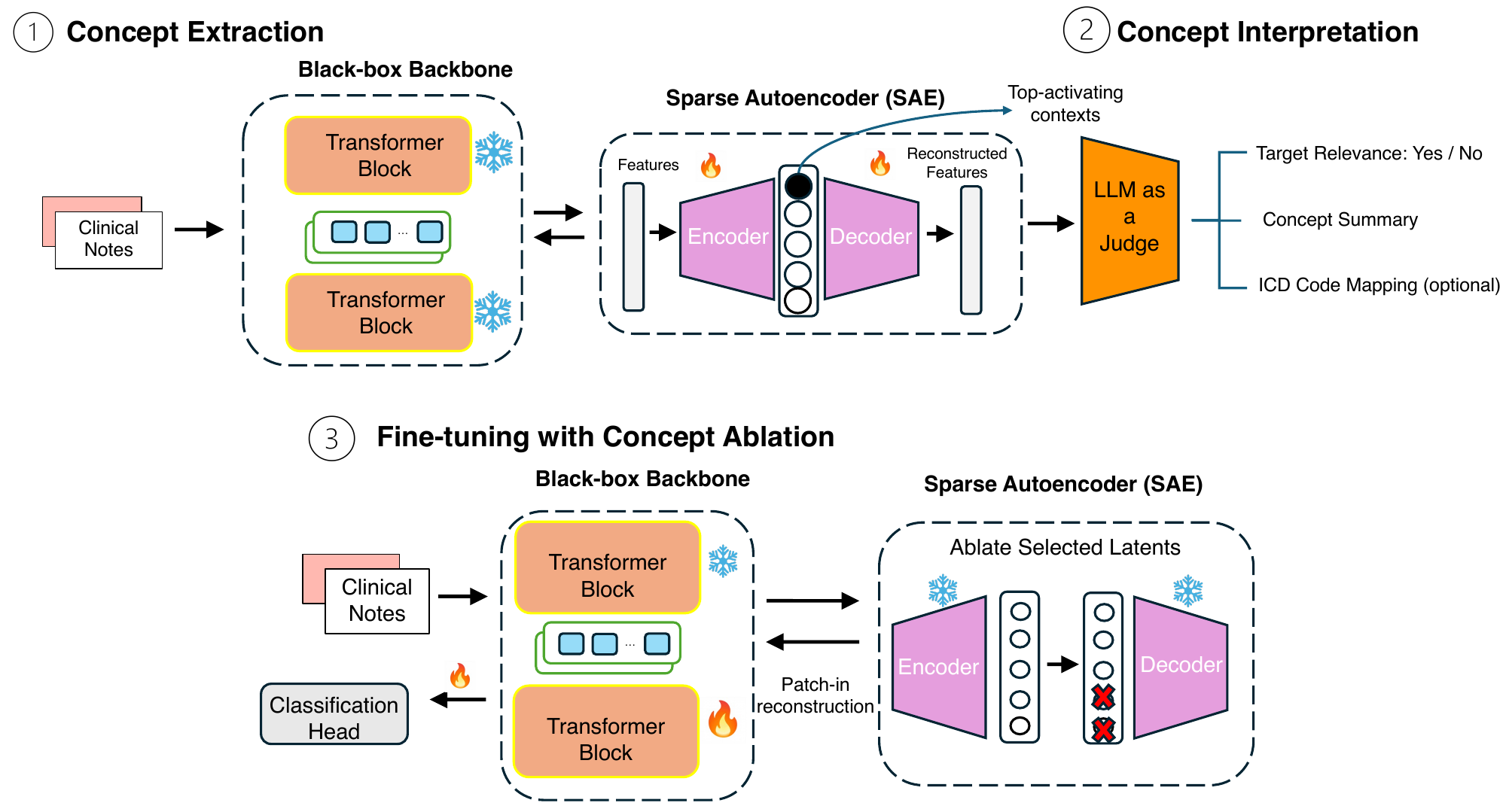} 
  \caption {\textbf{Framework for clinical concept extraction and controlled fine-tuning.} The pipeline consists of three stages: (1) \textbf{Concept Extraction} via a trained Sparse Autoencoder; (2) \textbf{Concept Interpretation} using an LLM to evaluate top activations; and (3) \textbf{Fine-tuning with Concept Ablation} to steer downstream model predictions.}
  \label{fig2}
\end{figure*}
We introduce an end-to-end framework that leverages Sparse Autoencoders to map internal representations to interpretable concepts and utilize them for controlled fine-tuning. The complete pipeline is illustrated in Figure \ref{fig2}.

\subsection{Concept Extraction via SAEs}
Let $f_\theta$ denote a pretrained Transformer encoder. Given an input clinical note
$x = (x_1,\ldots,x_T)$, we extract the token representations from layer $\ell$:
$$
H = f_\theta^{(\ell)}(x) \in \mathbb{R}^{T \times d}, \qquad h_t \in \mathbb{R}^d .
$$
where $h_t$ is the contextual hidden state of token $x_t$ at layer $\ell$. To convert these black-box features into an interpretable concept interface, we introduce a sparse concept decomposition module parameterized by an encoder-decoder pair ( $E_\phi, D_\psi$ ):
$$
z_t=E_\phi\left(h_t\right) \in \mathbb{R}^{d_\text{SAE}}, \quad \hat{h}_t=D_\psi\left(z_t\right) \in \mathbb{R}^d .
$$
The key requirement is that $z_t$ is sparse (only a few concept dimensions active per token), while $\hat{h}_t$ remains a faithful reconstruction of $h_t$. We therefore optimize a generic objective:
$$
\mathcal{L}_{\text {concept }}=\sum_{t=1}^T\left\|h_t-\hat{h}_t\right\|_2^2+\lambda \Omega(z)
$$
where $\Omega(\cdot)$ specifies the sparsity mechanism and can be instantiated in different SAE variants.

\subsection{Automated Concept Interpretation}
\label{sec:interp}
To make the learned concepts interpretable for auditing and steering, we employ an LLM-based interpretation pipeline. For each latent $j$, we extract a context window (e.g., $[t-w, t+w]$) around its top $k$ highest activating tokens $z_{t, j}$ from the clinical corpus. Prompted with these empirically grounded contexts, an LLM judge outputs three elements: (i) a concise description of the captured clinical concept, (ii) a binary classification of its relevance to the downstream task versus unintended artifacts (e.g., boilerplate), and (iii) a set of formal medical keywords suitable for querying standardized clinical databases. 

To prevent the LLM from hallucinating medical codes, we implement a retrieval-based safeguard: the generated keywords are used to query an ICD-10 database \citep{cdc_icd10cm}, allowing the LLM to assign verified clinical codes to the latent by selecting exclusively from the retrieved candidate set.

\subsection{Fine-Tuning with Concept Steering}
\label{sec:steering}

To operationalize the learned concepts, we integrate the frozen SAE as an
active steering intervention at its training layer $K$. We partition the
$L$-layer transformer into a frozen prefix (layers $1$ through $K$) and a
trainable suffix (layers $K{+}1$ through $L$). Given input tokens $x$, the
frozen prefix produces hidden states $h^{(K)}(x)$, which the SAE rewrites into
steered representations $\tilde{h}^{(K)}(x)$ as defined in
Equation~\eqref{eq:residual_correction} below. The trainable suffix then
processes the steered states.

Because clinical discharge notes routinely exceed the encoder's context window,
each document is partitioned into overlapping chunks. The suffix outputs are
reduced to a single chunk embedding by mean pooling over non-special tokens, and
chunk embeddings are combined into a document embedding through a learned
attention pool. The document embedding is then fed to a linear classification
head that produces the task logits.

\paragraph{Residual-correction intervention.}
Let $\mathcal{T}$ denote the set of task-relevant concept indices identified by
the LLM judge in Section~\ref{sec:interp}, and let
$\overline{\mathcal{T}}$ denote the indices labeled as task-irrelevant or
artifactual. Rather than replacing $h^{(K)}$ with the SAE reconstruction, which
would incur reconstruction error on signals the dictionary fails to capture, we
adopt a \emph{residual correction} that preserves the original hidden state and
subtracts only the contribution of the suppressed concepts:
\begin{equation}
\label{eq:residual_correction}
\tilde{h}^{(K)}_t
=
h^{(K)}_t
-
\sum_{j \in \overline{\mathcal{T}}}
z_{t,j} W_{\mathrm{dec}}[j,:],
\end{equation}
where $z_{t,j}$ is the SAE activation of latent $j$ at token $t$, and
$W_{\mathrm{dec}}[j,:]$ is its decoder direction. The unmodeled portion of
$h^{(K)}_t$, i.e., the SAE residual $h^{(K)}_t - D_\psi(z_t)$, passes through
unchanged, so clinical signal the SAE fails to reconstruct is preserved
verbatim.

\paragraph{Uniform artifact suppression.}
We subtract every latent the interpretation pipeline labels as
artifactual, regardless of its activation magnitude or estimated downstream
impact. Two considerations motivate this choice. First, even low-magnitude
artifact directions can correlate with institution- or template-specific
documentation patterns and shift the decision boundary under deployment
distribution shift. Second, the residual form in
Equation~\eqref{eq:residual_correction} makes uniform suppression more conservative: only the
explicit decoder contributions in $\overline{\mathcal{T}}$ are removed, while
the rest of $h^{(K)}_t$ flows through unchanged. This design deliberately
decouples \emph{what the model is forbidden to use}---any verified artifact,
set during training---from \emph{what evidence the model is shown to rely on},
which is defined as a post-hoc audit signal in Section~\ref{sec:attribution}.

\subsection{Per-Concept Attribution for Auditability}
\label{sec:attribution}

The suppression mechanism in Section~\ref{sec:steering} controls
\emph{what the model is forbidden to use}; clinical deployment additionally
requires evidence of \emph{what the trained model actually used}. We therefore
equip the framework with a post-hoc attribution module that, for every document,
scores each SAE latent by its contribution to the prediction without affecting
any trained parameter.

Let $F_\omega$ denote the deployed suffix classifier that maps the steered
layer-$K$ representation $\tilde{h}^{(K)}(x)$ to the positive-class logit,
\[
s(x) = F_\omega\!\left(\tilde{h}^{(K)}(x)\right).
\]
We use the logit rather than the probability to avoid vanishing gradients for
confident predictions. For latent $j$, we define the signed first-order
attribution as
\begin{equation}
\label{eq:attribution}
A_j(x)
=
\sum_{t=1}^{T}
z_{t,j}(x)
\left\langle
\nabla_t s(x),
W_{\mathrm{dec}}[j,:]
\right\rangle,
\end{equation}
where $\nabla_t s(x) \equiv
\partial s(x) / \partial \tilde{h}^{(K)}_t(x)$, and $z_{t,j}(x)$ is the SAE
activation of latent $j$ at token $t$.

Equation~\eqref{eq:attribution} is the first-order Taylor approximation of the
exact counterfactual logit change when concept $j$'s decoder direction is
removed from the steered state:
\begin{equation}
\label{eq:counterfactual}
\begin{aligned}
A_j(x)
\approx\;&
F_\omega\!\left(\tilde{h}^{(K)}(x)\right) \\
&-
F_\omega\!\left(
\tilde{h}^{(K)}(x)
-
z_{\cdot,j}(x) W_{\mathrm{dec}}[j,:]
\right).
\end{aligned}
\end{equation}
Positive values of $A_j(x)$ indicate concepts that push the prediction toward
mortality, whereas negative values indicate concepts that push against it. We
rank concepts by $|A_j(x)|$ for overall importance and by signed $A_j(x)$ when
presenting per-prediction evidence.

Computing Equation~\eqref{eq:attribution} requires one forward pass and one
backward pass per batch through $F_\omega$, plus a single matrix multiplication
against $W_{\mathrm{dec}}^\top$. This replaces the $|D|$ separate forward passes
required by exact counterfactual ablation, where $|D|$ is the SAE dictionary
size. In our implementation, the full MIMIC-IV test split is processed in
roughly six minutes on a single GPU. On $30$ held-out documents, the
per-document Spearman rank correlation between $A_j(x)$ and the exact
counterfactual effect is $\rho = 0.976$ with Pearson $r = 0.935$. Appendix
\ref{app:attribution} provides the closed-form expression for the linear case,
the validation protocol, and detailed use cases, including per-prediction
evidence trails, global auditing, and cost-efficient expert review.

\section{Results and Discussion}

\subsection{Dataset}
\label{sec:data}
We use the dataset introduced by  
\citet{yoon2025lcd} for 30-day out-of-hospital mortality prediction from ICU discharge notes in MIMIC-IV v2.2 \citep{PhysioNet-mimiciv-2.2}, with labels derived by linking encounters to an external death registry. The dataset contains 49,832 admissions/notes from 39,705 patients, with patient-level splits to avoid leakage and exclusions for in-hospital death and hospice disposition. Importantly, the classification target is extremely imbalanced (approximately 1,830 positives vs. 48,002 negatives for the 30-day task), posing a challenging setting for robust evaluation on long-form clinical text.

\subsection{Models and Baselines}
\label{sec:models_baselines}

For clinical note encoding, we use two domain-specific pretrained
Transformer backbones: ClinicalBERT \citep{alsentzer-etal-2019-publicly}
for inputs up to 512 tokens, and Clinical-Longformer
\citep{li2022clinical} for extended notes up to 4,096 tokens. Both are
continually pre-trained on the MIMIC corpus.

We compare CAST against standard fine-tuning of the same clinical encoders and
two zero-shot general-purpose LLM baselines, GPT-4 \citep{openai2024gpt4technicalreport} and
Llama-3-8B \citep{grattafiori2024llama}. We additionally include an
input-removal baseline that directly removes identified input-level artifacts
before fine-tuning, testing whether simple preprocessing alone is sufficient
to mitigate artifact-driven shortcuts. We also include two SAE-based
baselines: Self-Regul \citep{wu2025self}, which regularizes classification with
SAE-derived sparse features, and SAE-Probe, adapted from
\citet{gallifant-etal-2025-sparse}, which freezes the encoder and SAE, pools
SAE latent activations into document-level features, and trains a lightweight
classifier on top. Unlike CAST, both baselines use SAE latents as auxiliary
representations rather than as explicit artifact-suppression interventions
inside the fine-tuning loop.  Additional  details are provided in
Appendix~\ref{sec:baseline}.

\subsection{Training setup and hyperparameters} 
\paragraph{SAE pretraining.}
We train three variants of SAE, TopK, BatchTopK, and Matryoshka, on hidden activations from the frozen encoder over a mixed
clinical corpus of $200{,}000$ MIMIC notes; the corpus split is provided in
the Appendix~\ref{app:training}. We report TopK and Matryoshka in the main results, while BatchTopK results are provided in the Appendix~\ref{app:batchtopk-mimic4}. All SAEs use a dictionary size of
$d_{\mathrm{SAE}} = 8{,}192$ with $\mathrm{top}\text{-}k = 64$ active latents
per token. The Matryoshka variant uses nested group sizes
$\{512, 2048, 8192\}$, totaling $10{,}752$ latents. SAEs are trained with
AdamW using a learning rate of $1\mathrm{e}{-}3$ for approximately $410$M
tokens, and are frozen during all downstream fine-tuning.

\paragraph{Downstream training.}
We fine-tune all concept-guided models and fine-tuning baselines for $5$ epochs
using AdamW with differential learning rates of $5\mathrm{e}{-}5$ for the
backbone layers and $1\mathrm{e}{-}3$ for the classification head. The effective
batch size is $256$. To handle the approximately $27{:}1$ class imbalance, we
use class-weighted focal loss with $\gamma = 2.0$ and per-class weights
$\alpha = [1.0,\, N_{\mathrm{neg}} / N_{\mathrm{pos}}]$. We train each model for five epochs and evaluate the final checkpoint on the
held-out test set, reporting $F_1$ both at a fixed decision threshold of $0.5$
and at a validation-selected threshold $\tau^\star$ that maximizes $F_1$ on
the validation set. Additional optimization details are provided in
Appendix~\ref{app:training}.

\begin{table*}[!t]
\centering
\small
\setlength{\tabcolsep}{3.2pt}
\renewcommand{\arraystretch}{1.10}
\sisetup{detect-weight=true, detect-family=true}

\newcommand{\NA}{\multicolumn{1}{c}{--}}
\newcommand{\todo}{\multicolumn{1}{c}{\textit{TBD}}}
\newcommand{\yes}{\cmark}
\newcommand{\no}{\xmark}
\renewcommand{\best}[1]{\bfseries #1}

\newcommand{\casttext}[1]{\cellcolor{gray!12}#1}
\newcommand{\castcell}[1]{\cellcolor{gray!12}#1}
\newcommand{\castnum}[1]{\cellcolor{gray!12}#1}

\caption{
Performance on 30-day out-of-hospital mortality prediction from long discharge notes.
Zero-shot results are reported from \citet{yoon2025lcd}.
CAST denotes our concept-guided fine-tuning method. F1$_{\tau^\star}$ uses the decision threshold that maximizes F1 on the validation set.
}
\label{tab:performance}

\begin{adjustbox}{max width=\linewidth}
\begin{tabular}{@{}llllc
    S[table-format=1.4]
    S[table-format=1.4]
    S[table-format=1.4]
    S[table-format=1.4]
    S[table-format=1.4]
    S[table-format=1.4]
    S[table-format=1.4]
@{}}
\toprule
\textbf{Backbone} &
\textbf{Layer} &
\textbf{SAE} &
\textbf{Protocol} &
\textbf{Interp.} &
{\textbf{F1} $\uparrow$} &
{\textbf{F1}$_{\boldsymbol{\tau^\star}}$ $\uparrow$} &
{\textbf{AUROC} $\uparrow$} &
{\textbf{PR-AUC} $\uparrow$} &
{\textbf{Brier} $\downarrow$} &
{\textbf{NLL} $\downarrow$} &
{\textbf{ECE} $\downarrow$} \\
\midrule

GPT-4     & -- & -- & Zero-shot & \no & 0.3237 & \NA & \NA & \NA & \NA & \NA & \NA \\
Llama3-8B & -- & -- & Zero-shot & \no & 0.1948 & \NA & \NA & \NA & \NA & \NA & \NA \\

\midrule

\multirow{16}{*}{ClinicalBERT}
& \multirow{8}{*}{11}
& -- & Fine-tuning & \no
& 0.2602 & 0.2941 & 0.8320 & 0.2312 & 0.0908 & 0.3315 & 0.2246 \\

& & -- & Input removal & \no
& 0.2470 & \best{0.3002} & 0.8550 & 0.2450 & 0.1253 & 0.4228 & 0.2947 \\

& & \multirow{3}{*}{TopK}
& Self-Regul & \yes
& 0.1663 & 0.2415 & 0.8353 & 0.2011 & 0.1944 & 0.5756 & 0.3960 \\
& & & SAE-Probe & \yes
& 0.2632 & 0.2544 & 0.8265 & 0.2175 & 0.0806 & 0.3066 & 0.2078 \\
& & & \casttext{CAST} & \castcell{\yes}
& \castnum{\best{0.2961}}
& \castnum{0.3000}
& \castnum{\best{0.8579}}
& \castnum{\best{0.2460}}
& \castnum{0.0879}
& \castnum{0.3190}
& \castnum{0.2150} \\

& & \multirow{3}{*}{Matryoshka}
& Self-Regul & \yes
& 0.1631 & 0.2991 & 0.8390 & 0.2235 & 0.1948 & 0.5764 & 0.3963 \\
& & & SAE-Probe & \yes
& 0.2404 & 0.2227 & 0.8067 & 0.1743 & 0.0967 & 0.3348 & 0.2161 \\
& & & \casttext{CAST} & \castcell{\yes}
& \castnum{0.2749}
& \castnum{0.2927}
& \castnum{0.8382}
& \castnum{\best{0.2460}}
& \castnum{\best{0.0693}}
& \castnum{\best{0.2713}}
& \castnum{\best{0.1736}} \\

\cmidrule(lr){2-12}

& \multirow{8}{*}{8}
& -- & Fine-tuning & \no
& 0.2549 & 0.2612 & 0.8246 & 0.2089 & 0.0817 & 0.3129 & 0.2122 \\

& & -- & Input removal & \no
& 0.2240 & 0.2901 & 0.8230 & 0.1850 & 0.0911 & 0.3363 & 0.2234 \\

& & \multirow{3}{*}{TopK}
& Self-Regul & \yes
& 0.1639 & 0.2479 & 0.8351 & 0.2036 & 0.1949 & 0.5775 & 0.3979 \\
& & & SAE-Probe & \yes
& 0.2308 & 0.2345 & 0.7965 & 0.1643 & 0.0899 & 0.3318 & 0.2254 \\
& & & \casttext{CAST} & \castcell{\yes}
& \castnum{0.2837}
& \castnum{0.2699}
& \castnum{\best{0.8428}}
& \castnum{\best{0.2362}}
& \castnum{0.0837}
& \castnum{0.3168}
& \castnum{0.2163} \\

& & \multirow{3}{*}{Matryoshka}
& Self-Regul & \yes
& 0.1664 & 0.2594 & 0.8379 & 0.2059 & 0.1937 & 0.5746 & 0.3958 \\
& & & SAE-Probe & \yes
& 0.1637 & 0.1612 & 0.7709 & 0.1084 & 0.1086 & 0.3757 & 0.2554 \\    
& & & \casttext{CAST} & \castcell{\yes}
& \castnum{\best{0.2869}}
& \castnum{\best{0.2923}}
& \castnum{0.8422}
& \castnum{0.2355}
& \castnum{\best{0.0507}}
& \castnum{\best{0.2242}}
& \castnum{\best{0.1326}} \\

\midrule

\multirow{16}{*}{\makecell[l]{Clinical-\\Longformer}}
& \multirow{8}{*}{11}
& -- & Fine-tuning & \no
& 0.1759 & 0.3286 & 0.8666 & 0.2648 & 0.1909 & 0.5681 & 0.3908 \\

& & -- & Input removal & \no
& 0.2250 & 0.3242 & 0.8570 & 0.2650 & 0.1355 & 0.4315 & 0.2968 \\

& & \multirow{3}{*}{TopK}
& Self-Regul & \yes
& 0.1726 & 0.2771 & 0.8436 & 0.2342 & 0.1860 & 0.5575 & 0.3854 \\
& & & SAE-Probe & \yes
 & 0.1333 & 0.2313 & 0.8151 & 0.1553 & 0.2245 & 0.6350 & 0.3997 \\      
& & & \casttext{CAST} & \castcell{\yes}
& \castnum{\best{0.2794}}
& \castnum{\best{0.3412}}
& \castnum{\best{0.8740}}
& \castnum{\best{0.2702}}
& \castnum{\best{0.1217}}
& \castnum{\best{0.4097}}
& \castnum{\best{0.2882}} \\

& & \multirow{3}{*}{Matryoshka}
& Self-Regul & \yes
& 0.1943 & 0.2731 & 0.8465 & 0.2393 & 0.1760 & 0.5359 & 0.3723 \\
& & & SAE-Probe & \yes
 & 0.1454 & 0.1807 & 0.8028 & 0.1683 & 0.2014 & 0.5824 & 0.3853 \\
& & & \casttext{CAST} & \castcell{\yes}
& \castnum{0.2038}
& \castnum{0.2760}
& \castnum{0.8641}
& \castnum{0.2511}
& \castnum{0.1595}
& \castnum{0.4975}
& \castnum{0.3464} \\

\cmidrule(lr){2-12}

& \multirow{8}{*}{8}
& -- & Fine-tuning & \no
& 0.2028 & 0.2787 & \best{0.8556} & 0.2433 & 0.1506 & 0.4755 & 0.3289 \\

& & -- & Input removal & \no
& 0.3040 & 0.2843 & 0.8370 & 0.2300 & 0.1304 & 0.4427 & 0.3129 \\

& & \multirow{3}{*}{TopK}
& Self-Regul & \yes
& 0.1928 & 0.2775 & 0.8502 & 0.2470 & 0.1760 & 0.5365 & 0.3732 \\
& & & SAE-Probe & \yes
 &0.1777 & 0.1426 & 0.7782 & 0.1358 & 0.1160 & 0.3921 & 0.2652 \\               
& & & \casttext{CAST} & \castcell{\yes}
& \castnum{0.3041}
& \castnum{\best{0.3085}}
& \castnum{0.8523}
& \castnum{0.2360}
& \castnum{0.0900}
& \castnum{0.3396}
& \castnum{0.2375} \\

& & \multirow{3}{*}{Matryoshka}
& Self-Regul & \yes
& 0.1855 & 0.2741 & 0.8514 & 0.2506 & 0.1797 & 0.5443 & 0.3779 \\
& & & SAE-Probe & \yes
 & 0.1060 & 0.1815 & 0.7786 & 0.1095 & 0.2790 & 0.7566 & 0.4773 \\
& & & \casttext{CAST} & \castcell{\yes}
& \castnum{\best{0.3233}}
& \castnum{0.3073}
& \castnum{0.8531}
& \castnum{\best{0.2524}}
& \castnum{\best{0.0820}}
& \castnum{\best{0.3196}}
& \castnum{\best{0.2224}} \\

\bottomrule
\end{tabular}
\end{adjustbox}
\end{table*}

\subsection{Feature Extraction and Interpretation}
\label{sec:interp}

\begin{figure}[t]
  \centering 
  \includegraphics[width=\linewidth]{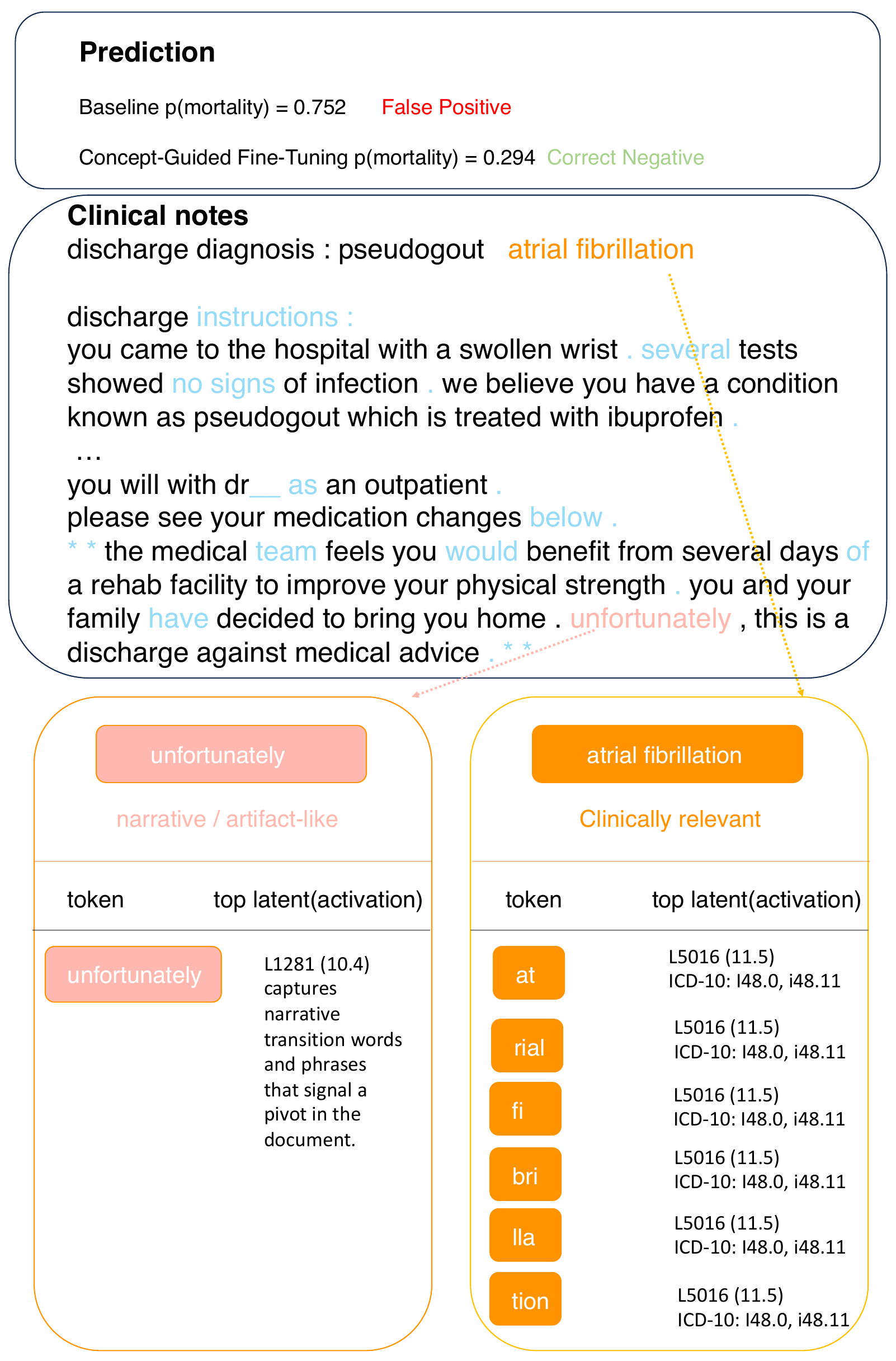} 
  \caption{Real clinical note case study: artifact suppression corrects a false positive while preserving clinical evidence.}
  \label{fig:cases}
\end{figure}

We train sparse autoencoders on frozen hidden activations from layer~$8$
(mid-encoder) and layer~$11$ (near the top of the encoder), and insert the
resulting modules back into the corresponding layers for downstream
fine-tuning (Section~\ref{sec:steering}).

To interpret the learned SAE latents, we apply an LLM-based labeling pipeline
to the maximally activating clinical contexts of each alive latent. For each
latent, \texttt{gemini-2.5-flash-lite}
\citep{gemini25report} is shown its top-10 activating contexts and asked
to produce a concise concept description, assign a semantic category, and determine whether the concept
is related to 30-day mortality. For diagnostic concepts, the LLM additionally
generates medical search terms that are used to retrieve candidate ICD-10-CM
codes; code assignment is restricted to the retrieved candidate set to reduce
code hallucination. Full prompts and implementation details are provided in
Appendix~\ref{app:interp_details}.

We use these interpretations to construct a conservative suppression set
$\overline{\mathcal{T}}$. Specifically, to improve labeling reliability, we run the interpretation pipeline
independently three times for each latent and include a latent in
$\overline{\mathcal{T}}$ only when all three runs classify it as either a
formatting or de-identification artifact and as not mortality-related.
This strict-consensus criterion is designed to reduce erroneous suppression
of clinically meaningful features. Across configurations, Fleiss' $\kappa$
\citep{Fleiss1971MeasuringNS} ranges from $0.707$--$0.749$ for mortality-related labels and
$0.749$--$0.806$ for artifact labels, while unanimous agreement ranges from
$82.6$--$87.7\%$ and $87.9$--$93.3\%$, respectively.
Per-configuration agreement statistics are reported in
Appendix~\ref{app:interp_details}.

At the aggregate level, the interpretation results show that many latents capture mortality-related concepts, while roughly half can be grounded to one
or more ICD-10 codes. ClinicalBERT yields a higher clinical-concept fraction
than Longformer, and layer-$11$ SAEs capture richer clinical semantics than
layer-$8$ SAEs. Per-configuration statistics are reported in
Appendix~\ref{app:latent_summary}.

At the instance level, Figure~\ref{fig:cases} illustrates how these
interpreted features translate into the steering behavior of CAST. In a
representative MIMIC-IV case, suppressing strict-consensus artifact latents
corrects a confident false-positive prediction while preserving activations
corresponding to genuine clinical evidence. 

\subsection{Performance Analysis}
\label{sec:performance}

Table~\ref{tab:performance} reports results on 30-day out-of-hospital
mortality prediction from long discharge notes. The zero-shot LLM baselines
provide two useful reference points: GPT-4 remains a strong closed-model
baseline, whereas Llama3-8B performs poorly on this long clinical-text task.
CAST instead uses compact clinical encoders and exposes predictions through
SAE-derived concepts, offering a more auditable alternative to general-purpose
LLMs.

Among encoder-based methods, the results highlight the importance of how SAE
information is used. Standard fine-tuning provides a strong but opaque
discriminative baseline, SAE-Probe uses SAE activations only as frozen
post-hoc features, and Self-Regul applies SAE-based regularization. CAST goes
beyond these alternatives by using SAE concepts as training-time steering
signals, yielding a stronger balance of discrimination, calibration, and
concept-level interpretability.

The input-removal baseline further tests whether direct preprocessing is
sufficient to address the identified artifacts. It is competitive on selected
metrics, particularly for ClinicalBERT at layer~$11$, confirming that visible
input-level artifacts can sometimes be removed effectively. However, its gains
are not consistent across backbones and layers. Input removal also does not provide the concept-level suppression
and audit trail available through CAST.

The pattern across backbones further suggests that CAST's gains are not tied to
a single encoder or SAE variant. For ClinicalBERT, TopK at layer~11 gives the
strongest discrimination, while Matryoshka at layer~8 yields the most reliable
risk estimates. For Clinical-Longformer, the layer~8 Matryoshka CAST model improves operating-point
performance while substantially reducing Brier score, NLL, and ECE. These
results suggest that SAE-guided steering can improve clinical risk prediction
without reducing the model to a purely post-hoc interpretability pipeline. We further assess CAST against the matched fine-tuning baselines using paired bootstrap tests with 10,000 resamples; full results are reported in Appendix~\ref{app:bootstrap}.

\subsection{Feature Importance}
\label{sec:feature_importance}

We use the per-concept attribution $A_j(x)$ defined in Section~\ref{sec:attribution} to perform a post-hoc analysis of the SAE latents that most influence mortality predictions. For each configuration,
we aggregate $A_j(x)$ over the MIMIC-IV test cohort by taking the signed mean
over documents in which latent $j$ is active. This produces a ranked set of SAE
concepts that most strongly influence the model's output.

Table~\ref{tab:top_concepts} reports the highest-magnitude signed attributions
for ClinicalBERT layer~$11$ across the three SAE variants. The positive
attributions correspond to clinically plausible mortality-related factors,
including palliative care, COPD exacerbation, metastatic disease, and acute
kidney failure. In contrast, the strongest negative attributions are associated
with recovery or lower-risk indicators, especially mobility and independent
ambulation. Notably, mobility-related concepts appear among the top negative
features for all three SAE variants in this layer, suggesting that the
attribution analysis captures stable clinically meaningful signals rather than
idiosyncratic features of a single SAE construction.

These findings complement the ablation results in
Section~\ref{sec:ablation}: suppressing mortality-related concepts harms
predictive performance, whereas suppressing artifact concepts improves it.
Appendix~\ref{app:clinical_concepts_detail} expands
Table~\ref{tab:top_concepts} with LLM-generated concept descriptions and
representative activating contexts, while
Appendix~\ref{app:artifact_examples} provides concrete examples of the
strict-consensus artifact set suppressed during training.
\begin{table}[t]
  \centering
  \small
  \setlength{\tabcolsep}{3.5pt}
  \renewcommand{\arraystretch}{1.12}

  \resizebox{\columnwidth}{!}{%
  \begin{tabular}{@{}llrl@{}}
    \toprule
    \textbf{SAE} & \textbf{Latent} & $\boldsymbol{\bar A_j}$ &
    \textbf{Concept (ICD-10-CM)} \\
    \midrule
    \multicolumn{4}{@{}l}{\emph{Push prediction $\uparrow$ mortality}} \\
    BatchTopK  & 1108 & $+0.18$ & palliative care (Z51.5) \\
    TopK       & 6730 & $+0.15$ & COPD exacerbation (J44.1) \\
    Matryoshka & 101  & $+0.04$ &
      \makecell[l]{metastatic disease /\\ secondary malignancy (C79.9)} \\
    TopK       & 915  & $+0.02$ & acute kidney failure (N17.9) \\
    \midrule
    \multicolumn{4}{@{}l}{\emph{Push prediction $\downarrow$ mortality}} \\
    BatchTopK  & 5888 & $-0.29$ & ambulate independently \\
    TopK       & 7211 & $-0.20$ & gait/mobility status (R26.89) \\
    TopK       & 6567 & $-0.19$ & independent ambulation \\
    BatchTopK  & 3189 & $-0.18$ & discharge mentions \\
    Matryoshka & 581  & $-0.13$ & gait/mobility status (R26.89) \\
    \bottomrule
  \end{tabular}%
  }

  \caption{
  Top signed-attribution SAE concepts for ClinicalBERT layer~$11$.
  Positive values push predictions toward mortality; negative values push them away.
  }
  \label{tab:top_concepts}
\end{table}

  \subsection{Ablation and Sensitivity Analysis}
\label{sec:ablation}
\begin{figure}[htbp]
  \centering 
\includegraphics[width=\linewidth]{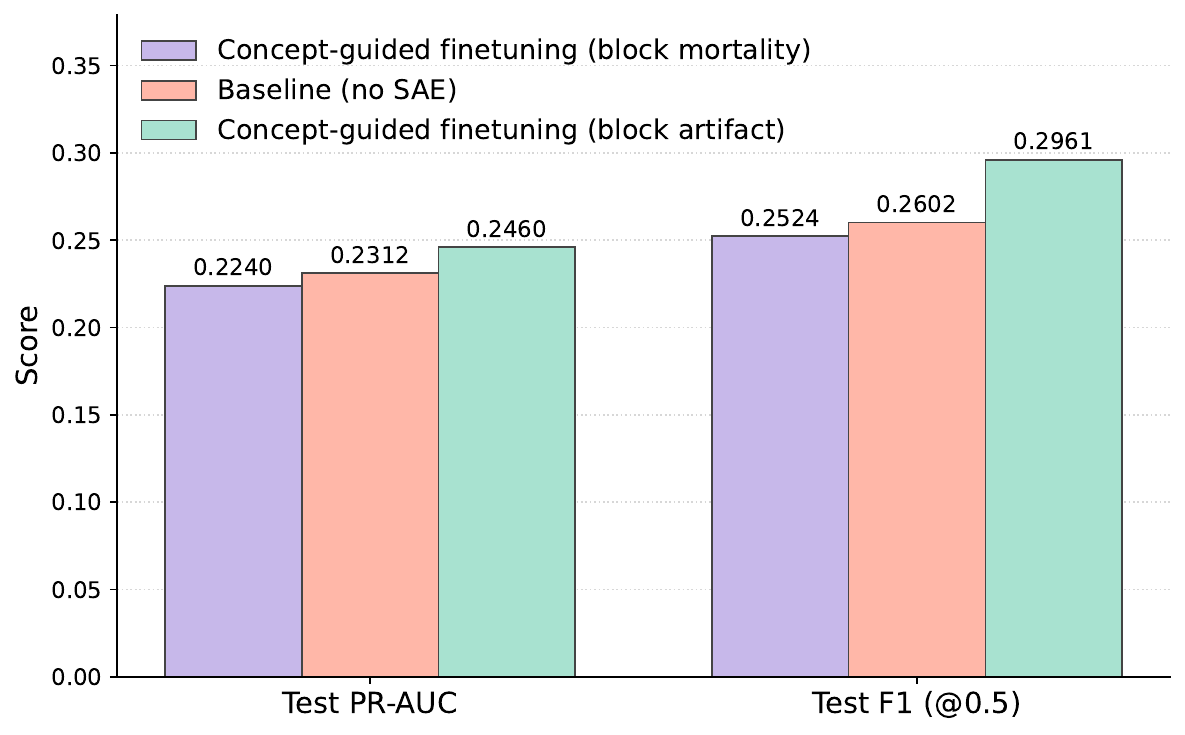} 
\caption{Ablation study of concept-guided fine-tuning. Blocking artifact-related concepts improves both PR-AUC and F1, while blocking mortality-related concepts degrades performance.}
  \label{fig:ablation}
\end{figure}

\paragraph{Concept ablation.} We conduct an ablation study on ClinicalBERT layer~$11$ to verify that CAST's
gains come from targeted concept-level intervention rather than simply adding an
SAE module. We compare three settings: standard fine-tuning, blocking
mortality-related concepts, and blocking artifact-related concepts. To make the
comparison controlled, we block the same number of mortality-related and
artifact-related latents.

As shown in Figure~\ref{fig:ablation}, matched-size interventions have opposite
effects: blocking mortality-related concepts hurts performance, while blocking
artifact-related concepts improves it. This suggests that CAST gains come from
targeted semantic steering rather than generic regularization, supporting SAE
latents as an actionable interface for auditing and control.

\paragraph{Layer and hyperparameter sensitivity.}
We additionally evaluate an earlier intervention layer (layer~4) for both backbones and vary the SAE dictionary size and TopK sparsity in a representative ClinicalBERT layer-11 configuration. Across these sensitivity analyses, CAST remains competitive over the tested settings. Full results are reported in Appendix~\ref{app:sensitivity}.

\section{Conclusion}

We introduced an SAE-guided fine-tuning framework for interpretable and
controllable clinical prediction. By turning sparse SAE concepts into
training-time steering signals, our method moves beyond post-hoc interpretation
and provides a direct interface for suppressing spurious artifacts. On MIMIC-IV
30-day mortality prediction, CAST maintains strong predictive performance while
improving calibration and exposing concept-level evidence for model decisions.
Overall, our framework offers a practical path toward auditable and reliable
clinical NLP systems.

\section*{Limitations}

Our evaluation focuses on a single prediction task using MIMIC-III and
MIMIC-IV discharge notes. Although these datasets span different time periods
and documentation systems, they originate from the same institution. Evaluation
on external institutions, additional note types, and other clinical tasks is
therefore an important direction for future work. Moreover, the modest absolute
F1 scores reflect the difficulty of this highly imbalanced task; we view CAST
as a research-stage auditing and steering framework rather than a deployable
clinical model. Establishing clinically useful operating points will require
prospective evaluation and clinician-in-the-loop assessment of the audit trail. Concept interpretation relies on an LLM judge and ICD-10-CM retrieval. Our
strict three-run consensus rule is designed to reduce erroneous suppression,
but clinician-annotated validation would provide a stronger assessment of
labeling reliability. Because artifact and workflow-related features may also
carry valid clinical or demographically correlated information, future work
should include expert review of suppression sets and subgroup-specific
evaluation before clinical use. Finally, CAST introduces additional offline cost for SAE pretraining and latent
interpretation, although the LLM is not used during test-time prediction.

\section*{Acknowledgment}
J. Mu and G. Chen's effort was partially supported by NSF grant DMS-2515263 and by the Patient-Centered Outcomes Research Institute (PCORI) Award ME-2024C1-37433. The statements in this work are solely the responsibility of the authors and do not necessarily represent the views of the Patient-Centered Outcomes Research Institute (PCORI), its Board of Governors, or the Methodology Committee.
\bibliography{custom}

\appendix
\clearpage
\section{Appendix}
\label{sec:appendix}

\subsection{Prompt Template} 
\begin{figure}[htbp]
\centering
\begin{tcolorbox}[
  colback=white,
  colframe=black,
  boxrule=0.5pt,
  sharp corners,
  left=5pt,
  right=5pt,
  top=5pt,
  bottom=5pt,
  width=\linewidth,
  breakable
]
\footnotesize
\textbf{System Prompt Template}\\
You are a clinical NLP expert analyzing features from a Sparse Autoencoder trained on Bio\_ClinicalBERT. The model processes clinical discharge notes and predicts 30-day out-of-hospital mortality.

\vspace{4pt}
\textbf{Concept Interpretation Prompt}\\
Analyze latent feature \#\{latent\_id\} from the SAE. Below are the top 10 tokens that maximally activate this latent, with activation strength and surrounding context from clinical discharge notes.

\vspace{2pt}
\texttt{\{examples\_block\}}

\vspace{2pt}
Based on these activation patterns, respond with ONLY a JSON object (no other text):

\vspace{2pt}
\begin{verbatim}
{
  "concept": "<1-2 sentence description>",
  "mortality_related": <true or false>,
  "mortality_reasoning": 
  "<1 sentence explanation>",
  "search_keywords": 
  ["<k1>", "<k2>", "<k3>", "<k4>", "<k5>"]
}
\end{verbatim}

\vspace{-4pt}
\textbf{Guidelines for \texttt{search\_keywords}:}
\begin{itemize}[leftmargin=*, itemsep=1pt, topsep=2pt]
  \item Provide exactly 5 formal medical terms suitable for searching an ICD-10-CM database.
  \item Use formal terminology, e.g., ``acute myocardial infarction'' instead of ``heart attack''.
  \item Include both specific condition terms and broader category terms.
  \item If the latent detects formatting or structural patterns, use the most relevant medical context visible in the examples.
\end{itemize}
\end{tcolorbox}

\caption{\textbf{Prompt template.} System prompt and user prompt used for concept interpretation.}
\label{fig:prompt-template}
\end{figure}

 \subsection{Per-Concept Attribution: Linear Limit, Validation, and Use Cases}                                                                                              
  \label{app:attribution}  
\paragraph{Closed-form linear case.}
When the suffix and prediction head collapse to a single linear map $W_g$
applied to a document-level aggregate
$\bar{z}(x) \in \mathbb{R}^{d_{\mathrm{SAE}}}$,
Equation~\eqref{eq:attribution} reduces to the analytic ablation expression
\[
A_j(x)
=
\bar{z}_j(x)
\left(
W_{\mathrm{dec}}[j,:] W_g^\top
\right),
\]
which can be evaluated without an additional network forward pass. The
gradient-based estimator in Equation~\eqref{eq:attribution} generalizes this
closed form to the non-linear deployed classifier.

\paragraph{Validation protocol.}
For each of $30$ randomly sampled test documents with predicted probability in
$[0.05, 0.95]$, we computed the exact logit-level counterfactual effect in
Equation~\eqref{eq:counterfactual} for the union of (i) the top-$100$ most
activated latents and (ii) the top-$100$ most attributive latents in that
document, yielding approximately $180$ concepts per document. We then computed
the per-document Spearman rank and Pearson correlations between $A_j(x)$ and
the exact effect. The resulting mean Spearman correlation is $\rho = 0.976$,
with median $0.978$, $Q_1 = 0.970$, and $Q_3 = 0.986$; the mean Pearson
correlation is $r = 0.935$. The mean relative $L_1$ error,
$|A_j - \Delta s_j| / \overline{|\Delta s_j|}$, is $0.38$. This residual is
composed of higher-order Taylor terms and does not substantially affect ranking
or sign.

\paragraph{Use cases.}
The attribution module serves three roles in our framework, none of which feeds
back into training. \textbf{(i) Per-prediction evidence trail.} For each
document, the top-ranked active concepts, together with their ICD-10 anchors,
form a human-readable explanation that clinicians can review alongside the risk
score (Figure~\ref{fig:cases}). \textbf{(ii) Global behavioral audit.}
Aggregating $A_j(x)$ over the test cohort reveals which clinical concepts
dominate the model's decisions and exposes any residual high-impact latent that
the LLM judge may have mislabeled or missed. \textbf{(iii) Cost-efficient
interpretation.} Because LLM-based labeling is rate-limited, ranking concepts by
$|A_j(x)|$ lets us focus expert review on the few hundred latents that most
influence predictions, rather than uniformly interpreting all $|D|$ dictionary
entries.

\subsection{LLM Interpretation Details}
\label{app:interp_details}

We label each alive SAE latent using \texttt{gemini-2.5-flash-lite}
\citep{gemini25report} through OpenRouter at temperature $0.3$. For each
latent, the LLM is shown the $10$ maximally activating contexts and asked to
summarize the captured concept in one or two sentences, assign one of seven
categories (\texttt{clinical}, \texttt{formatting\_artifact},
\texttt{deid\_artifact}, \texttt{temporal}, \texttt{demographic},
\texttt{structural\_marker}, or \texttt{other}), and decide whether the
concept is mortality-related. Diagnostic claims are grounded against a
retrieval lookup over the ICD-10-CM database to eliminate code hallucination.

We execute the interpretation pipeline three independent times per latent. A
latent enters the suppression set $\overline{\mathcal{T}}$ only when all three
runs label it as either \texttt{formatting\_artifact} or
\texttt{deid\_artifact} and not mortality-related. Table~\ref{tab:llm_agreement}
reports inter-run agreement for the two binary decisions used by CAST. Of the
latents labeled as artifacts in at least one run, only $44.3$--$51.2\%$ satisfy
the strict criterion, reflecting the precision-oriented design.

\begin{table}[htbp]
  \centering
  \small
  \setlength{\tabcolsep}{2.8pt}
  \renewcommand{\arraystretch}{1.06}
  \caption{LLM-labeling agreement across three independent runs. We report
  Fleiss' $\kappa$ and the fraction receiving a unanimous ($3$-of-$3$) label
  for mortality (Mort.) and artifact (Art.) decisions. Strict/any is the
  fraction of latents flagged as artifacts by at least one run that enter the
  strict-consensus suppression set.}
  \label{tab:llm_agreement}
  \resizebox{\columnwidth}{!}{%
  \begin{tabular}{@{}llccccc@{}}
    \toprule
    \textbf{Backbone} & \textbf{SAE / Layer} &
    $\boldsymbol{\kappa}_{\textbf{Mort.}}$ &
    $\boldsymbol{\kappa}_{\textbf{Art.}}$ &
    \textbf{3/3 Mort.} & \textbf{3/3 Art.} & \textbf{Strict/any} \\
    \midrule
    \multirow{6}{*}{ClinicalBERT}
      & BatchTopK / 11  & .749 & .796 & 87.7\% & 93.0\% & 47.5\% \\
      & Matryoshka / 11 & .748 & .793 & 87.2\% & 92.8\% & 48.7\% \\
      & TopK / 11       & .746 & .806 & 87.5\% & 93.3\% & 50.2\% \\
      & BatchTopK / 8   & .745 & .804 & 87.0\% & 93.1\% & 51.2\% \\
      & Matryoshka / 8  & .722 & .790 & 85.2\% & 91.9\% & 47.6\% \\
      & TopK / 8        & .733 & .801 & 86.3\% & 92.9\% & 48.0\% \\
    \midrule
    \multirow{6}{*}{\makecell[l]{Clinical-\\Longformer}}
      & BatchTopK / 11  & .726 & .759 & 83.7\% & 89.6\% & 45.7\% \\
      & Matryoshka / 11 & .717 & .749 & 82.6\% & 87.9\% & 44.3\% \\
      & TopK / 11       & .728 & .774 & 83.7\% & 90.0\% & 47.7\% \\
      & BatchTopK / 8   & .714 & .773 & 82.9\% & 89.8\% & 49.0\% \\
      & Matryoshka / 8  & .713 & .775 & 82.6\% & 89.6\% & 47.9\% \\
      & TopK / 8        & .707 & .767 & 82.7\% & 89.4\% & 47.5\% \\
    \bottomrule
  \end{tabular}%
  }
\end{table}

\subsection{Per-Configuration Interpretation Summary}
\label{app:latent_summary}

Table~\ref{tab:latent_summary} reports, for each backbone--SAE-type--layer
configuration, the number of LLM-interpreted latents, the fraction grounded to
ICD-10 codes, and the fraction classified as mortality-related. 

\paragraph{Backbone comparison.}
ClinicalBERT consistently yields a higher clinical-concept fraction than
Longformer at the same layer and SAE type. This disparity plausibly stems from
their distinct attention mechanisms: Longformer's local sliding-window attention
may limit the synthesis of scattered clinical entities, whereas ClinicalBERT's
full self-attention aggregates medical context into a denser semantic space,
yielding more interpretable latents.

\paragraph{Layer comparison.}
SAEs trained on layer~$11$ yield richer clinical semantics than those trained on
layer~$8$ across backbone--SAE combinations, reflecting the natural progression
of high-level, task-specific representations in deeper Transformer layers.

\paragraph{SAE variant comparison.}
Matryoshka SAEs match or slightly exceed TopK SAEs in clinical-concept fraction
within comparable settings. This pattern is consistent with the nested structure of
Matryoshka dictionaries, which encourages features to be organized across
coarser-to-finer representational groups.

The strict-consensus artifact set $\overline{\mathcal{T}}$ used in
Section~\ref{sec:steering} is a much smaller subset of the non-mortality side of
Table~\ref{tab:latent_summary}, requiring $3$-of-$3$ agreement across
independent LLM runs; see Section~\ref{sec:interp}.

\begin{table}[htbp]
  \centering
  \caption{
    Per-configuration summary of interpreted SAE latents. We report the
    percentage grounded to ICD-10 codes and the percentage classified as
    mortality-related across backbone architectures, SAE types, and layer depths.
  }
  \resizebox{\columnwidth}{!}{%
    \begin{tabular}{@{}llSS@{}}
      \toprule
      \textbf{Backbone}
      & {\makecell{\textbf{SAE type}\\\textbf{/ Layer}}}
      & {\makecell{\textbf{ICD-10}\\\textbf{(\%)}}}
      & {\makecell{\textbf{Mortality}\\\textbf{(\%)}}} \\
      \midrule
      \multirow{6}{*}{ClinicalBERT}
      & TopK / 8        & 55.3 & 78.3 \\
      & TopK / 11       & 57.6 & 79.2 \\
      & BatchTopK / 8   & 55.5 & 78.4 \\
      & BatchTopK / 11  & 57.5 & 79.9 \\
      & Matryoshka / 8  & 55.5 & 77.4 \\
      & Matryoshka / 11 & 57.7 & 78.5 \\
      \midrule
      \multirow{6}{*}{Longformer}
      & TopK / 8        & 50.2 & 72.9 \\
      & TopK / 11       & 50.9 & 72.7 \\
      & BatchTopK / 8   & 51.2 & 72.2 \\
      & BatchTopK / 11  & 51.1 & 72.6 \\
      & Matryoshka / 8  & 49.7 & 71.7 \\
      & Matryoshka / 11 & 48.1 & 71.1 \\
      \bottomrule
    \end{tabular}%
  }
  
  \label{tab:latent_summary}
\end{table}

\subsection{Training Details} 
 \label{app:training}
\paragraph{SAE pretraining corpus.}
We pretrain SAEs on $200{,}000$ MIMIC notes: $120{,}000$ from MIMIC-III
narrative records and $80{,}000$ from MIMIC-IV discharge summaries.

\paragraph{SAE optimization.}
Each SAE is trained for $100{,}000$ steps with $4{,}096$ tokens per step,
corresponding to approximately $410$M training tokens. We use AdamW with
learning rate $1\mathrm{e}{-}3$, $\ell_1$ coefficient $5\mathrm{e}{-}4$, and an
auxiliary penalty of $1.56\mathrm{e}{-}2$ for latents inactive for more than
$1{,}000$ batches.

\paragraph{Downstream fine-tuning.}
We use AdamW with weight decay $0.01$ and an effective batch size of $256$,
using gradient accumulation when needed. All reported runs use random seed $42$.

\subsection{Layer and Hyperparameter Sensitivity}
\label{app:sensitivity}

We extend the layer analysis to layer~$4$ and conduct controlled sensitivity
experiments on ClinicalBERT with a TopK SAE at layer~$11$. Unless otherwise varied, the
default setting is $d_{\mathrm{SAE}}=8{,}192$, $k=64$, learning rate
$5\times10^{-5}$, and five fine-tuning epochs.

\paragraph{Earlier-layer intervention.}
Table~\ref{tab:layer4_sensitivity} complements the layer-$8$ and layer-$11$
results in Table~\ref{tab:performance}. Early-layer behavior is more dependent
on the SAE architecture: TopK performs best for ClinicalBERT, whereas
Matryoshka performs best for Clinical-Longformer.

\begin{table}[htbp]
  \centering
  \small
  \setlength{\tabcolsep}{3.0pt}
  \renewcommand{\arraystretch}{1.08}
  \caption{Layer-$4$ results on MIMIC-IV, complementing the layer-$8$ and
  layer-$11$ evaluations in Table~\ref{tab:performance}.}
  \label{tab:layer4_sensitivity}
  \resizebox{\columnwidth}{!}{%
  \begin{tabular}{@{}llccc@{}}
    \toprule
    \textbf{Backbone} & \textbf{Method} & \textbf{F1} & \textbf{AUROC} & \textbf{PR-AUC} \\
    \midrule
    \multirow{3}{*}{ClinicalBERT}
      & Fine-tuning       & .2935 & .8415 & .2429 \\
      & TopK CAST         & \textbf{.3061} & .8511 & \textbf{.2574} \\
      & Matryoshka CAST   & .2306 & \textbf{.8538} & .2342 \\
    \midrule
    \multirow{3}{*}{\makecell[l]{Clinical-\\Longformer}}
      & Fine-tuning       & .2746 & .8464 & .2378 \\
      & TopK CAST         & .2432 & \textbf{.8570} & .2485 \\
      & Matryoshka CAST   & \textbf{.3025} & .8484 & \textbf{.2526} \\
    \bottomrule
  \end{tabular}%
  }
\end{table}

\paragraph{SAE dictionary size and sparsity.}
Table~\ref{tab:sae_sensitivity} varies $d_{\mathrm{SAE}}$ and $k$. Performance remains stable across the tested settings, suggesting that CAST is not sensitive to a particular dictionary size or sparsity level.

\begin{table}[htbp]
  \centering
  \small
  \setlength{\tabcolsep}{4.5pt}
  \renewcommand{\arraystretch}{1.06}
  \caption{Dictionary-size and TopK-sparsity sensitivity for ClinicalBERT
  layer~$11$ CAST.}
  \label{tab:sae_sensitivity}
  \begin{tabular}{@{}rrccc@{}}
    \toprule
    $\boldsymbol{d_{\mathrm{SAE}}}$ & $\boldsymbol{k}$ & \textbf{F1} & \textbf{AUROC} & \textbf{PR-AUC} \\
    \midrule
     4,096 &  64 & .2703 & .8511 & \textbf{.2531} \\
     8,192 &  32 & .2569 & .8525 & .2530 \\
     8,192 &  64 & \textbf{.2961} & \textbf{.8579} & .2460 \\
     8,192 & 128 & .2801 & .8502 & .2344 \\
    16,384 &  64 & .2881 & .8538 & .2504 \\
    \bottomrule
  \end{tabular}
\end{table}

\subsection{Baseline implementations.}
\label{sec:baseline}
For fair comparison with the protocols,
we adapt both SAE-based baselines to our long-text setting while preserving
their characteristic designs.

\paragraph{Self-Regul} \citep{wu2025self} is implemented as a linear probe on a
frozen document embedding:
\begin{equation}
\mathbf{x}
=
\mathbf{h}
-
\mathbf{W}_{\text{dec}}^\top
\left(
\mathrm{ReLU}\!\left(
\mathbf{W}_{\text{enc}}^\top
(\mathbf{h}-\mathbf{b}_{\text{dec}})
\right)
\odot
\mathbb{1}_{\overline{\mathcal{T}}}
\right),
\end{equation}
where $\mathbf{h}$ is obtained by attention-masked mean pooling within chunks,
followed by mean pooling across chunks, and
$\overline{\mathcal{T}}$ is the same strict-consensus artifact set used by
CAST. We retain the original weight-projection penalty,
\begin{equation}
\lambda
\left\lVert
\mathbf{W}_{\text{enc}}^\top
(\mathbf{w}-\mathbf{b}_{\text{dec}})
[\overline{\mathcal{T}}]
\right\rVert_{1},
\end{equation}
and select $\lambda$ on the development set.

\paragraph{SAE-Probe} \cite{gallifant-etal-2025-sparse} sums per-token SAE activations
across the entire document, excluding special tokens, optionally binarizes at
the paper-default threshold $\tau=1$, and trains a linear classifier on the
resulting $d_{\text{SAE}}$-dimensional feature vector. Because
$\mathbf{F}=\sum_t \mathbf{f}_t$ saturates on long discharge notes
($\sim$5{,}000 tokens $\times$ $k=64$ active latents per token), we drop the
binarization step for SAE configurations whose summed features exceed $95\%$
density on the training set and keep $\tau=1$ otherwise.

For both baselines and CAST, we use the same single-layer classifier head,
class-weighted focal loss
($\gamma=2$,
$\boldsymbol{\alpha}=[1,\,N_{\text{neg}}/N_{\text{pos}}]$),
AdamW optimization, effective batch size $256$, and $5$ training epochs. The learning rate is
$1\!\times\!10^{-3}$ for the classifier head and $5\!\times\!10^{-5}$ for any
trainable backbone layers. Self-Regul and SAE-Probe keep both                                                    
  the encoder and the SAE frozen; CAST and the standard fine-tuning baseline                                                    
  tune the upper backbone layers under an identical schedule.                                                                   

\subsection{Paired Bootstrap Analysis}
\label{app:bootstrap}

We draw
$10{,}000$ resamples from the shared test set and apply the same resample to
each TopK or Matryoshka CAST model and its matched fine-tuning baseline. Tests are one-sided in the
direction of improvement: $\Delta=\mathrm{CAST}-\mathrm{FineTune}>0$ for F1,
F1$_{\tau^\star}$, AUROC, and PR-AUC, and $\Delta<0$ for Brier score, NLL, and
ECE. Table~\ref{tab:bootstrap_effects} reports the  $p$-value for all
 metrics in Table~\ref{tab:performance}.
\begin{table*}[htbp]
  \centering
  \small
  \setlength{\tabcolsep}{5.0pt}
  \renewcommand{\arraystretch}{1.10}
  \caption{One-sided paired-bootstrap $p$-values for TopK and Matryoshka CAST
  relative to the matched fine-tuning baseline ($B=10{,}000$). The alternative is an
  improvement by CAST. Bold entries denote  $p<.05$.}
  \label{tab:bootstrap_effects}
  \resizebox{\textwidth}{!}{%
  \begin{tabular}{@{}lllccccccc@{}}
    \toprule
    \textbf{Backbone} & \textbf{Layer} & \textbf{SAE} & \textbf{F1} &
    \textbf{F1$_{\tau^\star}$} & \textbf{AUROC} & \textbf{PR-AUC} &
    \textbf{Brier} & \textbf{NLL} & \textbf{ECE} \\
    \midrule
    ClinicalBERT & 11 & TopK & $\mathbf{.0107}$ & $.3389$ &
      $\mathbf{.0002}$ & $.1504$ & $\mathbf{<.001}$ &
      $\mathbf{<.001}$ & $\mathbf{<.001}$ \\
    & 11 & Matryoshka & $.1769$ & $.5458$ & $.2269$ & $.1030$ &
      $\mathbf{<.001}$ & $\mathbf{<.001}$ & $\mathbf{<.001}$ \\
    ClinicalBERT & 8 & TopK & $.1499$ & $.4766$ &
      $\mathbf{.0108}$ & $.1705$ & $.9982$ & $.9925$ & $1.0000$ \\
    & 8 & Matryoshka & $.2950$ & $.1011$ &
      $\mathbf{.0330}$ & $.4500$ & $\mathbf{<.001}$ &
      $\mathbf{<.001}$ & $\mathbf{<.001}$ \\
    \makecell[l]{Clinical-\\Longformer} & 11 & TopK & $\mathbf{<.001}$ & $.2211$ &
      $\mathbf{.0086}$ & $.2335$ & $\mathbf{<.001}$ &
      $\mathbf{<.001}$ & $\mathbf{<.001}$ \\
    & 11 & Matryoshka & $\mathbf{<.001}$ & $.9978$ & $.9983$ & $.9783$ &
      $\mathbf{<.001}$ & $\mathbf{<.001}$ & $\mathbf{<.001}$ \\
    \makecell[l]{Clinical-\\Longformer} & 8 & TopK & $\mathbf{<.001}$ & $.0815$ &
      $.6877$ & $.7398$ & $\mathbf{<.001}$ &
      $\mathbf{<.001}$ & $\mathbf{<.001}$ \\
    & 8 & Matryoshka & $\mathbf{<.001}$ & $.1095$ & $.6131$ & $.1962$ &
      $\mathbf{<.001}$ & $\mathbf{<.001}$ & $\mathbf{<.001}$ \\
    \bottomrule
  \end{tabular}%
  }
\end{table*}

\subsection{Additional BatchTopK SAE Results on MIMIC-IV}
\label{app:batchtopk-mimic4}

Table~\ref{tab:batchtopk_mimic4} reports BatchTopK SAE performance on the
MIMIC-IV discharge-note held-out test set across the four backbone--layer
configurations, alongside standard fine-tuning, Self-Regul, and SAE-Probe
baselines. We omit BatchTopK from the main results table
(Table~\ref{tab:performance}) for space, since its overall behavior closely tracks
that of the TopK SAE. The trends mirror those in the main table.

\subsection{Evaluation on MIMIC-III}
MIMIC-III is a publicly available critical-care
database covering ICU admissions at Beth Israel Deaconess Medical Center
between 2001 and 2012. We use its discharge notes, paired with the same
30-day out-of-hospital mortality label as MIMIC-IV
(Section~\ref{sec:data}), giving $n=42{,}548$ notes with a positive rate of
$4.15\%$. The two corpora share the source institution, broad discharge-note
format, and ICD-based outcome coding, but differ in documentation period,
note templates, attending-physician populations, and EHR version. We
therefore use MIMIC-III as a second naturally heterogeneous evaluation set,
rather than as a distribution-shift  test.

Table~\ref{tab:performance_ood} summarizes the resulting performance. Across all  backbone--layer settings, the best-performing CAST variant achieves higher F1 than the corresponding fine-tuning baseline. The largest absolute
gains occur on Clinical-Longformer, reaching up to $+0.11$
F\textsubscript{1} at layer~8 with the Matryoshka SAE, suggesting that the
fine-tuned Longformer model is more sensitive to corpus-specific surface
patterns. The calibration metrics show the most consistent improvement: the
best-calibrated model in every cell is a CAST variant, with Brier, NLL, and
ECE substantially reduced across all four backbone-layer settings.

The relative ranking of SAE families also carries over from
Table~\ref{tab:performance}: Matryoshka is preferred for ClinicalBERT, whereas TopK
and BatchTopK are preferred for Clinical-Longformer.  Overall, the calibration and F\textsubscript{1} gains
of artifact-guided suppression transfer cleanly from MIMIC-IV to MIMIC-III,
supporting our claim that CAST recovers more clinically stable mortality
signals than its fine-tuned counterparts.

\begin{table*}[!t]
\centering
\small
\setlength{\tabcolsep}{3.2pt}
\renewcommand{\arraystretch}{1.10}
\sisetup{detect-weight=true, detect-family=true}

\newcommand{\NA}{\multicolumn{1}{c}{--}}
\newcommand{\todo}{\multicolumn{1}{c}{\textit{TBD}}}
\newcommand{\yes}{\cmark}
\newcommand{\no}{\xmark}

\newcommand{\casttext}[1]{\cellcolor{gray!12}#1}
\newcommand{\castcell}[1]{\cellcolor{gray!12}#1}
\newcommand{\castnum}[1]{\cellcolor{gray!12}#1}

\caption{
Performance of the BatchTopK SAE on 30-day out-of-hospital                   
  mortality prediction on the \textbf{MIMIC-IV} discharge-note test set                                                         
  ($n=7{,}568$, positive rate $\sim$3.6\%). Models are trained on the                                                           
  MIMIC-IV training split following the protocol of                                                                             
  Section~\ref{sec:data}. F\textsubscript{1} is computed at a                                                       
  fixed decision threshold of $0.5$.
}
\label{tab:batchtopk_mimic4}

\begin{adjustbox}{max width=\linewidth}
\begin{tabular}{@{}llllc
    S[table-format=1.4]
    S[table-format=1.4]
    S[table-format=1.4]
    S[table-format=1.4]
    S[table-format=1.4]
    S[table-format=1.4]
@{}}
\toprule
\textbf{Backbone} &
\textbf{Layer} &
\textbf{SAE} &
\textbf{Protocol} &
\textbf{Interp.} &
{\textbf{F1} $\uparrow$} &
{\textbf{AUROC} $\uparrow$} &
{\textbf{PR-AUC} $\uparrow$} &
{\textbf{Brier} $\downarrow$} &
{\textbf{NLL} $\downarrow$} &
{\textbf{ECE} $\downarrow$} \\
\midrule

\multirow{8}{*}{ClinicalBERT}
& \multirow{4}{*}{11}
& -- & Fine-tuning & \no
& 0.2602 & 0.8320 & 0.2312 & 0.0908 & 0.3315 & 0.2246 \\
& & \multirow{3}{*}{BatchTopK}
& Self-Regul & \yes
& 0.1617 & 0.8358 & 0.2109 & 0.1954 & 0.5776 & 0.3969 \\
& & & SAE-Probe & \yes
& 0.1288 & 0.7282 & 0.0786 & 0.1738 & 0.5158 & 0.2670 \\
& & & \casttext{CAST} & \castcell{\yes}
& \castnum{\best{0.2769}}
& \castnum{\best{0.8428}}
& \castnum{\best{0.2473}}
& \castnum{\best{0.0825}}
& \castnum{\best{0.3034}}
& \castnum{\best{0.2004}} \\

\cmidrule(lr){2-11}

& \multirow{4}{*}{8}
& -- & Fine-tuning & \no
& 0.2549 & 0.8246 & 0.2089 & 0.0817 & 0.3129 & 0.2122 \\
& & \multirow{3}{*}{BatchTopK}
& Self-Regul & \yes
& 0.1676 & 0.8384 & 0.2110 & 0.1930 & 0.5732 & 0.3951 \\
& & & SAE-Probe & \yes
& 0.1419 & 0.7339 & 0.0832 & 0.0984 & 0.4943 & 0.1099 \\
& & & \casttext{CAST} & \castcell{\yes}
& \castnum{\best{0.2744}}
& \castnum{\best{0.8411}}
& \castnum{\best{0.2188}}
& \castnum{\best{0.0525}}
& \castnum{\best{0.2274}}
& \castnum{\best{0.1344}} \\

\midrule

\multirow{8}{*}{\makecell[l]{Clinical-\\Longformer}}
& \multirow{4}{*}{11}
& -- & Fine-tuning & \no
& 0.1759 & \best{0.8666} & 0.2648 & 0.1909 & 0.5681 & 0.3908 \\
& & \multirow{3}{*}{BatchTopK}
& Self-Regul & \yes
& 0.1804 & 0.8433 & 0.2363 & 0.1820 & 0.5489 & 0.3802 \\
& & & SAE-Probe & \yes
 & 0.0985 & 0.6671 & 0.0585 & 0.1739 & 0.5341 & 0.3683 \\ 
& & & \casttext{CAST} & \castcell{\yes}
& \castnum{\best{0.1991}}
& \castnum{0.8649}
& \castnum{\best{0.2652}}
& \castnum{\best{0.1627}}
& \castnum{\best{0.5013}}
& \castnum{\best{0.3472}} \\

\cmidrule(lr){2-11}

& \multirow{4}{*}{8}
& -- & Fine-tuning & \no
& 0.2028 & \best{0.8556} & 0.2433& 0.1506 & 0.4755 & 0.3289 \\
& & \multirow{3}{*}{BatchTopK}
& Self-Regul & \yes
& 0.1905 & 0.8498 & \best{0.2470}  & 0.1763 & 0.5371 & 0.3735 \\
& & & SAE-Probe & \yes
 & 0.1981 & 0.8013 & 0.1496 & 0.1160 & 0.3889 & 0.2652         \\
& & & \casttext{CAST} & \castcell{\yes}
& \castnum{\best{0.2927}}
& \castnum{0.8527}
& \castnum{0.2351}
& \castnum{\best{0.0897}}
& \castnum{\best{0.3355}}
& \castnum{\best{0.2336}} \\

\bottomrule
\end{tabular}
\end{adjustbox}
\end{table*}

\begin{table*}[!t]
  \centering
  \small
  \setlength{\tabcolsep}{3.2pt}
  \renewcommand{\arraystretch}{1.10}
  \sisetup{detect-weight=true, detect-family=true}

  \newcommand{\yes}{\cmark}
  \newcommand{\no}{\xmark}

  \caption{
    Model Performance on MIMIC-III discharge notes
    ($n=42{,}548$, positive rate $4.15\%$). Models are trained on the
    MIMIC-IV train set and evaluated on MIMIC-III without further
    tuning. CAST denotes our concept-guided fine-tuning method.
  }
  \label{tab:performance_ood}

  \begin{adjustbox}{max width=\linewidth}
  \begin{tabular}{@{}llllc
    S[table-format=1.4]
    S[table-format=1.4]
    S[table-format=1.4]
    S[table-format=1.4]
    S[table-format=1.4]
    S[table-format=1.4]
  @{}}
    \toprule
    \textbf{Backbone}
      & \textbf{Layer}
      & \textbf{Protocol}
      & \textbf{SAE}
      & \textbf{Interp.}
      & {\textbf{F1} $\uparrow$}
      & {\textbf{AUROC} $\uparrow$}
      & {\textbf{PR-AUC} $\uparrow$}
      & {\textbf{Brier} $\downarrow$}
      & {\textbf{NLL} $\downarrow$}
      & {\textbf{ECE} $\downarrow$} \\
    \midrule

    \multirow{8}{*}{ClinicalBERT}
      & \multirow{4}{*}{11}
      & Fine-tuning & --         & \no
      & 0.2969 & 0.8260 & 0.2631 & 0.1019 & 0.3628 & 0.2424 \\
      & & CAST        & TopK       & \yes
      & 0.2933 & \best{0.8415} & \best{0.2737} & 0.1054 & 0.3693 & 0.2472 \\
      & & CAST        & BatchTopK  & \yes
      & 0.2965 & 0.8318 & 0.2668 & 0.1016 & 0.3583 & 0.2369 \\
      & & CAST        & Matryoshka & \yes
      & \best{0.3130} & 0.8317 & 0.2688 & \best{0.0802} & \best{0.3066} & \best{0.1948} \\
    \cmidrule(lr){2-11}

      & \multirow{4}{*}{8}
      & Fine-tuning & --         & \no
      & 0.2968 & 0.8033 & 0.2517 & 0.0938 & 0.3488 & 0.2330 \\
      & & CAST        & TopK       & \yes
      & 0.3037 & \best{0.8275} & \best{0.2697} & 0.0957 & 0.3498 & 0.2341 \\
      & & CAST        & BatchTopK  & \yes
      & 0.2992 & 0.8141 & 0.2449 & 0.0640 & 0.2643 & 0.1564 \\
      & & CAST        & Matryoshka & \yes
      & \best{0.3213} & 0.8137 & 0.2639 & \best{0.0604} & \best{0.2558} & \best{0.1497} \\
    \midrule

    \multirow{8}{*}{\makecell[l]{Clinical-\\Longformer}}
      & \multirow{4}{*}{11}
      & Fine-tuning & --         & \no
      & 0.2334 & 0.8484 & 0.2786 & 0.1713 & 0.5271 & 0.3617 \\
      & & CAST        & TopK       & \yes
      & \best{0.3176} & \best{0.8532} & \best{0.2805} & \best{0.1250} & \best{0.4237} & \best{0.2930} \\
      & & CAST        & BatchTopK  & \yes
      & 0.2280 & 0.8487 & 0.2804 & 0.1690 & 0.5192 & 0.3550 \\
      & & CAST        & Matryoshka & \yes
      & 0.2790 & 0.8287 & 0.2220 & 0.1309 & 0.4372 & 0.3004 \\
    \cmidrule(lr){2-11}

      & \multirow{4}{*}{8}
      & Fine-tuning & --         & \no
      & 0.2216 & \best{0.8504} & 0.2771 & 0.1630 & 0.5040 & 0.3423 \\
      & & CAST        & TopK       & \yes
      & 0.3233 & 0.8427 & 0.2696 & 0.0970 & 0.3572 & 0.2428 \\
      & & CAST        & BatchTopK  & \yes
      & 0.3305 & 0.8465 & \best{0.2784} & 0.0946 & 0.3482 & 0.2355 \\
      & & CAST        & Matryoshka & \yes
      & \best{0.3308} & 0.8428 & 0.2743 & \best{0.0861} & \best{0.3301} & \best{0.2215} \\
    \bottomrule
  \end{tabular}
  \end{adjustbox}
\end{table*}

\subsection{Top Clinical Concepts by Attribution}
\label{app:clinical_concepts_detail}

Table~\ref{tab:appendix_clinical} expands Table~\ref{tab:top_concepts} with
the LLM-generated concept description and a representative maximally activating
context for each of the top signed-attribution clinical concepts on
ClinicalBERT layer~$11$ across the three SAE variants. The activating contexts
are sampled directly from MIMIC-IV discharge notes; the symbol \texttt{/}
denotes a line break, originally represented as \texttt{<cr>}.

\begin{table*}[t]
  \centering
    \caption{
    Top signed-attribution clinical concepts $\bar{A}_j$ on ClinicalBERT
    layer~$11$. Concepts are aggregated across the three SAE variants and shown
    in descending order of $|\bar{A}_j|$ within each direction. Concept
    descriptions are produced by the retrieval-grounded LLM judge
    (Section~\ref{sec:interp}); contexts are sampled from the SAE training
    corpus.
  }
  \resizebox{\textwidth}{!}{%
    \begin{tabular}{@{}c l r l p{0.32\textwidth} p{0.22\textwidth}@{}}
      \toprule
      \textbf{SAE} &
      \textbf{L\#} &
      {$\boldsymbol{\bar{A}_j}$} &
      \textbf{ICD-10} &
      \textbf{Concept (LLM description)} &
      \textbf{Representative activating context} \\
      \midrule
      \multicolumn{6}{@{}l}{\emph{Push prediction $\uparrow$ mortality}} \\
      \midrule
      TopK
      & 4415
      & $+0.21$
      & R19.1
      & Physical exam finding of a ``soft'' abdomen, often with descriptors of normal bowel sounds.
      & \texttt{/ abd : soft , nt , nd , normal} \\
      BatchTopK
      & 1108
      & $+0.18$
      & Z51.5
      & Palliative-care services, often in the context of serious illness or cancer treatment.
      & \texttt{second / round of palliative chemo} \\
      TopK
      & 6730
      & $+0.15$
      & G70.01, J44.1
      & Worsening of existing symptoms or emergence of new symptoms.
      & \texttt{temp > 101.5, worsening pain , / drainage} \\
      BatchTopK
      & 7197
      & $+0.12$
      & I27.23, J60--J70
      & Mentions of the lung, particularly lung cancer, lung masses, and other pulmonary pathology.
      & \texttt{lung sounds very diminished l lung fie...} \\
      TopK
      & 5454
      & $+0.13$
      & F41.1, F06.4
      & Anxiety, often in the context of psychiatric conditions such as depression.
      & \texttt{20mg daily for depression / anxiety} \\
      TopK
      & 3294
      & $+0.09$
      & --
      & Improvement of clinical conditions, indicating a positive response to treatment.
      & \texttt{area on buttocks has improved since admission} \\
      \midrule
      \multicolumn{6}{@{}l}{\emph{Push prediction $\downarrow$ mortality}} \\
      \midrule
      BatchTopK
      & 5888
      & $-0.29$
      & --
      & Patient's ability to ambulate independently, indicating a high level of functional status.
      & \texttt{activity status : ambulatory - independent} \\
      BatchTopK
      & 7211
      & $-0.26$
      & --
      & Mobility status, specifically ambulatory and independent.
      & \texttt{/ activity status : ambulatory - independent} \\
      TopK
      & 6567
      & $-0.19$
      & R26.89, M62.3
      & Independent ambulation status, indicating self-mobility ability.
      & \texttt{> activity status : ambulatory - independent} \\
      TopK
      & 1645
      & $-0.12$
      & Y83.8, Y83
      & Presence or absence of major surgical or invasive procedures.
      & \texttt{major surgical or invasive procedure :} \\
      BatchTopK
      & 7643
      & $-0.11$
      & R41.82
      & Presence and integrity of various bodily systems, particularly neurocognitive functions.
      & \texttt{attention / concentration : grossly intact} \\
      TopK
      & 3082
      & $-0.07$
      & R41.82, I69.0
      & Use of the word ``intact'' when describing the status of bodily functions.
      & \texttt{attention / concentration : grossly intact} \\
      \bottomrule
    \end{tabular}%
  }

  \label{tab:appendix_clinical}
\end{table*}

\subsection{Examples of Suppressed Artifacts}
\label{app:artifact_examples}

To make the strict-consensus suppression set $\overline{\mathcal{T}}$ concrete,
Table~\ref{tab:appendix_artifacts} shows representative entries from the 
strict-consensus artifact latents identified on the ClinicalBERT layer~$11$
TopK SAE. Two broad sub-categories dominate: \emph{formatting artifacts},
including line-break tokens, list-item delimiters, parenthetical notations, and
structural section markers , and \emph{de-identification
artifacts}, including placeholder tokens introduced by the MIMIC de-identification
pipeline such as \texttt{[**...**]} . These patterns may correlate
with patient outcomes at the dataset level, for example through documentation
density or note structure, but they do not directly reflect patient physiology.
Their decoder contributions are therefore subtracted from $\tilde{h}^{(K)}$
during fine-tuning (Section~\ref{sec:steering}).

\begin{table*}[t]
  \centering 
  \caption{
    Representative entries from the strict-consensus artifact latents on
    the ClinicalBERT layer~$11$ TopK SAE. The $3$-of-$3$ LLM agreement rule
    (Section~\ref{sec:interp}) classifies these latents as documentation-format
    or de-identification artifacts unrelated to patient state; their decoder
    contributions are subtracted from $\tilde{h}^{(K)}$ during fine-tuning.
  }
  \resizebox{\textwidth}{!}{%
    \begin{tabular}{@{}l c p{0.42\textwidth} p{0.28\textwidth}@{}}
      \toprule
      \textbf{Sub-category} &
      \textbf{L\#} &
      \textbf{LLM concept description} &
      \textbf{Representative activating context} \\
      \midrule
      formatting
      & 7488
      & Line breaks and formatting elements, especially the \texttt{<cr>} token in lists.
      & \texttt{please take all pills ) / / followup instructions} \\
      formatting
      & 4310
      & Structural transition from \emph{response} to \emph{plan} sections in clinical notes.
      & \texttt{cr > response : / plan : / chronic o} \\
      formatting
      & 1625
      & Parenthetical confirmation notations, e.g., \texttt{(x)} yes or \texttt{( )} no.
      & \texttt{services / considered ? ( x ) yes - ( ) no} \\
      formatting
      & 8123
      & List-item or code-snippet delimiters, especially the semicolon character.
      & \texttt{care , asking app questions . a ; / loving p ; c} \\
      formatting
      & 2201
      & Phrases indicating that specific details or instructions are provided elsewhere.
      & \texttt{of systems is unchanged from admission except as n} \\
      \midrule
      de-id
      & 2195
      & De-identified first- or last-name tokens of the form \texttt{[**FirstName5(NamePattern1) X**]}.
      & \texttt{needed . [ * * first name5 ( namepattern1 ) 1875 *} \\
      de-id
      & 7360
      & De-identified date placeholders of the form \texttt{[**YYYY-MM-DD**]}.
      & \texttt{l / [ * * 2189 - 4 - 8 * * ]} \\
      de-id
      & 6789
      & De-identified patient-name or title placeholders, e.g., ``Dr. X spoke with \ldots''.
      & \texttt{[ * * last name ( stitle ) 1430 * * ] spoke} \\
      \bottomrule
    \end{tabular}%
  }

  \label{tab:appendix_artifacts}
\end{table*}

\subsection{Artifacts and licensing}
We use MIMIC-III and MIMIC-IV under the PhysioNet credentialed access and data use requirements, and only for retrospective research on clinical NLP. 
We use publicly available pretrained clinical encoders in a manner consistent with their research use for clinical text modeling, and use ICD-10-CM terminology only for retrieval-based concept grounding. 
We do not redistribute MIMIC clinical notes, and any released code or trained artifacts will exclude patient text and follow the corresponding dataset and model usage terms. 
The models and artifacts created in this work are intended only for research and auditing of clinical language models, not for deployment in clinical decision-making without further validation, approval, and compliance review.
\end{document}